\pdfoutput=1

\documentclass[11pt]{article}
\usepackage[preprint]{acl}

\usepackage{times}
\usepackage{latexsym}

\usepackage{algorithm}    
\usepackage{algpseudocode}
\usepackage{amssymb}
\usepackage{amsmath}
\usepackage{listings}
\usepackage{booktabs}

\newcommand{\cpm}[1]{\textcolor{gray}{$_{\pm #1}$}}

\usepackage{float}
\newfloat{algorithm}{t}{lop}
\usepackage[T1]{fontenc}

\usepackage[utf8]{inputenc}

\usepackage{microtype}

\usepackage{inconsolata}

\usepackage{graphicx}

\usepackage{textcomp}

\usepackage{booktabs}
\usepackage{multirow} 
\usepackage{amsmath}

\usepackage{graphicx} 
\usepackage{multirow} 
\usepackage{booktabs} 

\title{PLD+: Accelerating LLM inference by leveraging Language Model Artifacts}

\author{Shwetha Somasundaram, Anirudh Phukan, Apoorv Saxena \\
        Adobe Research, India
\\  \{ shsomasu, phukan, apoorvs\}@adobe.com}

\begin{document}
\maketitle
\begin{abstract}
 To reduce the latency associated with autoretrogressive LLM inference, speculative decoding has emerged as a novel decoding paradigm, where future tokens are drafted and verified in parallel. However, the practical deployment of speculative decoding is hindered by its requirements for additional computational resources and fine-tuning, which limits its out-of-the-box usability. To address these challenges, we present PLD+, a suite of novel algorithms developed to accelerate the inference process of LLMs, particularly for input-guided tasks. These tasks, which include code editing, text editing, summarization, etc., often feature outputs with substantial overlap with their inputs—an attribute PLD+ is designed to exploit. PLD+ also leverages the artifacts (attention and hidden states) generated during inference to accelerate inference speed. We test our approach on five input-guided tasks and through extensive experiments we find that PLD+ outperforms all tuning-free approaches. In the greedy setting, it even outperforms the state-of-the-art tuning-dependent approach EAGLE on four of the tasks. (by a margin of upto 2.31 in terms of avg. speedup). Our approach is tuning free, does not require any additional compute and can easily be used for accelerating inference of any LLM.
\end{abstract}

\section{Introduction}
Large language models have emerged as the foundational building blocks for a wide array of user-facing applications, enabling unprecedented capabilities in natural language processing and generation \cite{liu2023summary}. 
However, the autoregressive decoding approach employed by these large language models introduces significant inference latency, a challenge that becomes more severe as the model size and generation length increase \cite{xia2024unlocking}. This latency can pose a barrier to the integration of these models into interactive applications, underscoring the importance of developing efficient decoding strategies to address this fundamental limitation.

One strategy that has been proposed to mitigate the inference latency challenge faced by large language models is the Speculative Decoding paradigm, which operates based on the Draft and Verify principle to accelerate the inference process \cite{stern2018blockwise, leviathan2023fast}. The two key steps in this paradigm are (a) efficient generation of multiple future tokens in the drafting step and (b) parallel verification of the drafted tokens using the target Language Model to ensure quality and alignment.

Classic drafting strategies usually either employ a smaller independent model to efficiently draft tokens or leverage the target LLM itself, utilizing techniques such as incorporating additional FFN heads \cite{stern2018blockwise, cai2024medusa} or layer skipping \cite{zhang2023draft}. However, these methods often require extensive tuning, which needs to be performed for every new model, and can be time and resource-intensive.
\begin{figure*}[htbp]
    \centering
    \includegraphics[width=\textwidth]{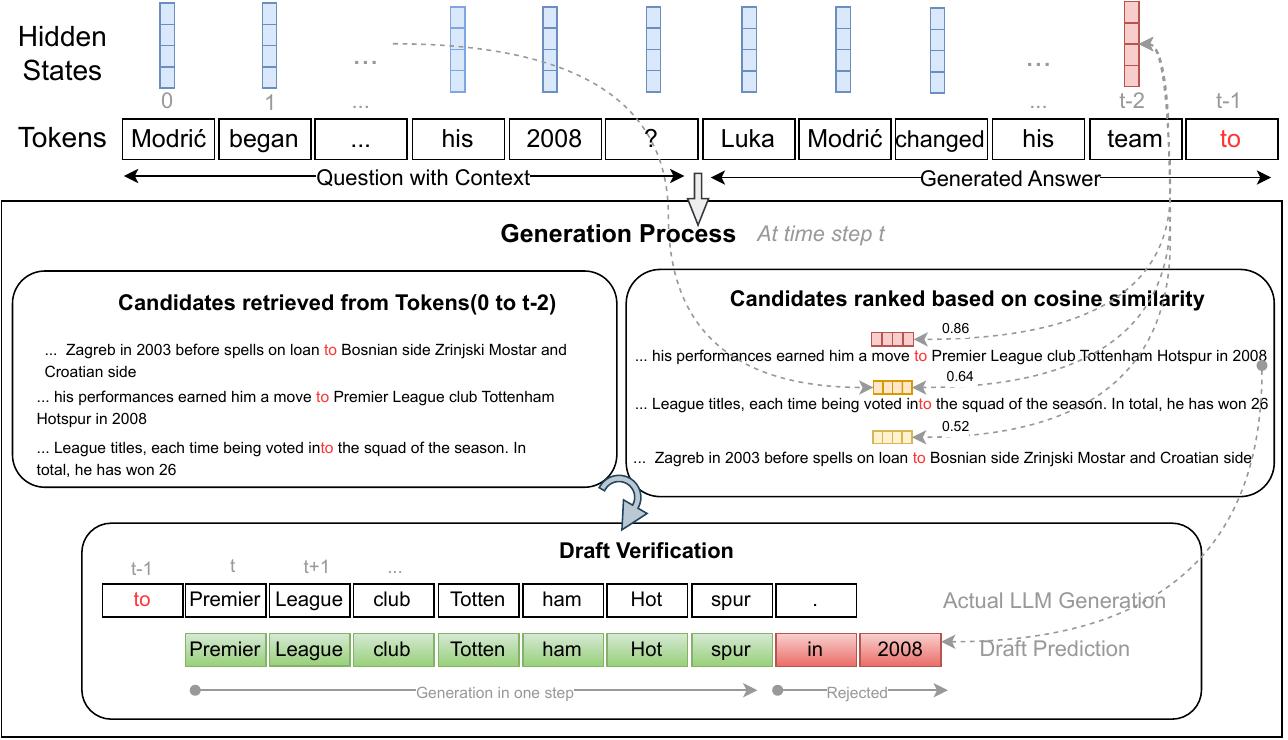}
    \caption{Overview of PLD+. During the generation of token $t$, possible drafts are retrieved from the context by searching for the same tokens as token $t-1$ ("to"). These candidates are then ranked using the information present in the model artifacts (hidden states in the figure) corresponding to token $t-2$. The text following "to" in the highest ranked candidate is proposed as the draft. The draft is verified against the actual LLM generation and all successfully verified tokens are generated in time step $t$, resulting in speedup.}
    \label{fig:overview}
\end{figure*}

In contrast, the PLD/LLMA approach gets rid of the need for any additional draft model by simply selecting text spans from the input as drafts, thus being extremely simple and demonstrating speedup on "input-guided tasks" \cite{saxena2023pld, yang2023inference}. Input-guided tasks are those with context-rich prompts, where the output is directly informed by or closely aligned with the input information, such as summarization, retrieval-augmented generation, and code/text editing. The simplicity and effective nature of PLD has resulted in its integration with the \texttt{transformers} library, underlining the importance of a plug-and-play method that achieves speedup on "input-guided tasks". 

In this paper, we build upon PLD and propose Prompt Lookup Decoding+ (PLD+), which leverages the information present in the model artifacts (attentions and hidden states generated during inference) to generate better drafts. In essence, we propose a tuning-free plug-and-play speculative decoding technique that leverages the model artifacts computed during the generation process. We outline the key contributions of our paper below:

\begin{itemize}
    \item We introduce PLD+, a suite of plug and play tuning-free speculative decoding algorithms that exploit the overlap between the input and the output generations. PLD+  leverages model artifacts to intelligently select draft spans and accelerate LLM inference. 
    \item The strengths of our method are that it does
not require any finetuning, it can be easily
applied to any LLM and it relies solely on
model artifacts that are computed during the generation process.
    \item Through extensive experiments, we show that PLD+ outperforms tuning-free baselines for both greedy decoding and sampling and that it can sometimes even outperform the best tuning-dependent approach EAGLE. PLD+ outperforms tuning-free baselines across model families as well.
   
\end{itemize}

\section{Related Work}

Speculative Decoding, introduced by \citet{stern2018blockwise}, represents a novel paradigm aimed at enhancing the efficiency of LLM inference. Instead of generating tokens sequentially, Speculative Decoding drafts multiple future tokens efficiently and then verifies them in parallel.

Drafting strategies are distinctly categorized into Independent and Self-Drafting methodologies. Independent Drafting leverages external models for the drafting process. This method often necessitates special training or fine-tuning to align with the target LLM's output characteristics. Examples include the use of smaller models within the same model family to draft tokens for their larger counterparts  \cite{chen2023accelerating, leviathan2023fast}, capitalizing on inherent architectural and training similarities to boost alignment and reduce additional training requirements.

PLD/LLMA \cite{saxena2023pld, yang2023predictive} is a simple yet highly effective independent drafting strategy for speeding up LLM inference in "input-guided tasks". It leverages string matching to generate candidates, capitalizing on the n-gram overlap between input and output. By replacing external models with string matching, this approach benefits from being tuning-free and model agnostic.

Self-Drafting strategies employ the target LLM itself for drafting, introducing innovative methods like appending multiple \texttt{[PAD]} tokens to simulate future tokens \cite{monea2023pass, santilli2023accelerating} or employing early exiting \cite{yang2023inference} and layer skipping techniques within the LLM \cite{zhang2023draft} to expedite the drafting phase. These approaches minimize the overhead associated with integrating separate drafter models and ensure closer inherent alignment with the target LLM's output patterns.

Substantial research efforts have focused on improving speculative decoding, yet the specific and ubiquitous setting of "input-guided tasks" has seen limited advancement beyond the simple PLD method. REST \cite{he2023rest} while similar in flavor to PLD, retrieves drafts from an external datastore to avoid being limited to a small context. The authors of REST note that for best results, they require a large datastore constructed using LLM generations itself, resulting in a considerable overhead. 
Our approach improves PLD by ranking the retrieved candidates from the input context based on semantics (rather than heuristics) using the model artifacts. Also by focusing on "input guided tasks", we get rid of the additional overhead of an external datastore required by REST.

\section{Background: Autoregressive decoding and Speculative Decoding}
\label{sec:background}
\textbf{Autoregressive Decoding}: 
Given an input sequence $x=x_1,x_2,\dots x_t$, an autoregressive language model $\mathcal{M_{\text{q}}}$ generates the next token $x_{t+1}$ as, $x_{t+1} \sim q_{t+1} = \mathcal{M_{\text{q}}}(x|x_1,x_2,...,x_t)$, 
where $q_{t+1}$ is the conditional probability calculated by the model $\mathcal{M_{\text{q}}}$ over its vocabulary $\mathcal{V}$. Based on the sampling scheme, the token $x_{t+1}$ is sampled from the probability distribution. 
The causal dependency of each decoding step on the results of previous decoding steps makes autoregressive decoding slow and inefficient due to its inability to fully utilize the parallelization capability of GPUs.

\label{sec:autoreg}

\textbf{Speculative Decoding}: Speculative Decoding \cite{stern2018blockwise,leviathan2023fast}, can be expressed as a Draft and Verify paradigm \cite{xia2024unlocking}. 
For a given decoding step $t$, a drafting algorithm  speculates $\textsc{K}$ draft tokens $\hat{x_1},\dots,\hat{x_{\textsc{K}}}$ efficiently, which are then verified by the language model $\mathcal{M_{\text{q}}}$. In the standard speculative decoding approach, the drafting algorithm used is a smaller language model $\mathcal{M_{\text{p}}}$.

In the verification phase, the draft tokens $\hat{x_1},\dots,\hat{x_{\textsc{K}}}$ are verified by the model $\mathcal{M_{\text{q}}}$ using a single forward pass. Given the input sequence $x=x_1,x_2,\dots x_t$ and the draft tokens $\hat{x_1},\dots,\hat{x_{\textsc{K}}}$, the model computes $\textsc{K}+1$ probability distributions at once.
Each drafted token is verified as per a verification strategy. Common verification criteria include rejection resampling \cite{leviathan2023fast,chen2023accelerating} and greedy acceptance \cite{stern2018blockwise}. Only the draft tokens that meet the verification strategy are retained in order to maintain consistency of generation with respect to standard autoregressive decoding using the model $\mathcal{M_{\text{q}}}$.

\section{Our approach: PLD+}
\label{sec:our_approach}

In the following sections, we explain the drafting \S\ref{sec:pldplusdraft} and verification \S\ref{sec:pldplusverify} algorithms of our approach. Figure \ref{fig:overview} provides an overview of the generation process using PLD+ for inference. 

\subsection{Notations}
 Formally, given an input sequence $x=x_1,x_2,... x_t$ passed to a language model $\mathcal{M_{\text{q}}}$ for decoding step $t$, our approach predicts and verifies draft tokens $\hat{x_1},\dots,\hat{x_{K}}$ leveraging model attentions $\mathbf{A}$ or model hidden states $\mathbf{H}$ computed during the generation process. The value \textsc{K} denotes the number of draft tokens that are predicted.  The hidden states $\mathbf{H}$ is a vector with dimensions $(L, |x_{<t}|, d)$ and  the model attention states $\mathbf{A}$ is a vector with dimensions $(L, G, |x_{<t}|, |x_{<t}|)$, where, $L$ is the number of layers in the model,
$G$ is the number of attention heads per layer, $|x_{<t}|$ is the sequence length of the input tokens before decoding step $t$, $d$ is the embedding size.

 \subsection{Drafting Algorithm}
 \label{sec:pldplusdraft}
 Our goal is to select an optimal token span from $x$ such that we exploit the overlap between the input sequence and the generation sequence. 
 
 To achieve this, we first identify a set of positions $P$, where the last generated token $x_{t}$ occurs in the input token sequence $x$.
 \begin{equation}  
    P = \{ j \mid x_j = x_t, j < t \}  
\end{equation} 
  
 To maximize inference speedup, we need to \textbf{rank} these occurrences intelligently. We hypothesize that the artifacts computed during the generation process captures contextual information which can be utilized to choose the optimal occurrence. In the following two sections, we describe how model hidden states and attentions are used to find the best occurrence $j^*$, for accelerated inference.
 
\subsubsection{ Ranking occurrences using model attentions}
\label{sec:att_rank}

The model attentions $\mathbf{A}$ computed by the model across different layers $l$ and heads $g$ is available for the sequence $x_{<t}$. The most straightforward method of ranking occurrences is to aggregate all attention maps across $l$ and $g$ for token $x_{t-1}$ using max or sum operation and choose the position $j^*$ which has the highest value. Recent  work in mechanistic interpretability \cite{olsson2022context,bansal-etal-2023-rethinking} indicate that  there exists induction heads which drive the in context learning ability of models. Induction heads are defined as attention heads that engage in two specific behaviors: prefix matching, which involves locating a previous instance of the current token within the context, and copying, which entails duplicating the sequence that follows the identified token. The behavior of induction heads can be highly useful for accelerating inference for input-guided tasks.  

\textbf{Identification of relevant attention heads: } We identified heads that can be relevant by first generating the outputs $o_t \in O$ and attentions $\mathbf{A}$ for a set of prompts. For each generated token $o_{t}$ present in the input $x$, we find the set of positions $P$, from where the token could have been "copied", i.e, positions where the generated token is present in the input. We then choose the position $r^*$ which has the maximum overlapping suffix with the generated output. We then iterate over all of the attention heads $L \times G$ and keep track of the attention heads where the token position $x_{r^*}$ has the maximum attention. We repeat this process for every prompt and for a given model $\mathcal{M_{\text{q}}}$ we identify relevant attention heads , $G^*$ from layers $L^*$. 

After identifying the relevant attention heads, we aggregate the attention scores from the heads in $G^*$ across layers $L^*$ using the max operation, and then we select the position $j^*$ that has the highest value.

\begin{equation}  
\label{eq:att_max}  
    j^* = \underset{j \in P}{\mathrm{argmax}} \, \left( \max_{l \in L^*, g \in G^*} \mathbf{A}^{(l,g)}_{t-1} \right)  
\end{equation}



 \subsubsection{ Ranking occurrences using hidden states} The hidden states $\mathbf{H}$ computed by the model across different layers $l$ is available for the sequence $x_{<t}$. The hidden states corresponding to the last generated token $x_{t}$ is not available. Therefore, for each position $j \in P$, we compute the cosine similarity between $\mathbf{H}^{(l)}_{j-1}$ and $\mathbf{H}^{(l)}_{t-1}$ and select the occurrence $j^*$ which results in the highest similarity value. 
 \begin{equation}  
\label{eq:hidden_state_computation}
    j^* = \underset{ j \in P}{\mathrm{argmax}} \, \mathrm{cos\_sim}(\mathbf{H}^{(l)}_{j-1}, \mathbf{H}^{(l)}_{t-1})  
   \end{equation}

\subsection{Draft Prediction} We predict $\hat{x_i}, \dots , \hat{x}_{i+\textsc{K}}$ as the future tokens, where $\textsc{K}$ denotes the number of predicted draft tokens. The number of draft tokens \textsc{K}, the layer $l$ and the head $g$ are the hyper parameters. 
\begin{equation*}
     \textit{draft tokens} \quad \hat{x_i}=x_{j^*+i} , i=1, \dots\textsc{K}
\end{equation*}

\subsection{Verification Algorithm}
\label{sec:pldplusverify}
The goal of the verification phase is to ensure that the tokens generated by PLD+ are the same as those generated by standard autoregressive decoding. To achieve this, we first pass the input sequence $x$ along with the draft tokens $\hat{x_i}$ to obtain conditional probabilities for future positions ($t+i,i= 1, \dots K$) using $\mathcal{M_{\text{q}}}$. Using these probabilities, we sample new tokens at the future positions. We verify if the sampled tokens match with the draft tokens at each position. After the first mismatch we discard the subsequent draft tokens. 

\subsection{Application Scenarios}
\label{sec:applications}

Our motivation is to accelerate generation in tasks where the generation outputs have significant overlaps with the input context. As enumerated by \citet{yang2023inference}: Multi-turn conversation, Retrieval Augmented Generation and Cache assisted Generation naturally fall under the input-guided tasks paradigm. We conduct our experiments on a broader range of tasks: code editing, text editing (short), text editing (long), multiturn conversation, and summarization.

\begin{table*}[t!]
\small 
\centering
\resizebox{\linewidth}{!}{
\begin{tabular}{@{}ll|ccccc|c@{}}
\toprule
\multicolumn{2}{l|}{\bf Methods} & \bf Summarization & \bf Code Editing & \bf Text Editing (Short) & \bf Text Editing (Long) &\begin{tabular}[c]{@{}c@{}}\bf Multi-turn\\\bf Conversation\end{tabular} &\begin{tabular}[c]{@{}c@{}}\bf Avg. Throughput\\\bf (\#tokens/s)\end{tabular} \\
\midrule
\multirow{11}{*}{\rotatebox{90}{\texttt{Vicuna-7B}}} 
&Autoregressive Decoding                    & 1.00$\times$\cpm{0.00} & 1.00$\times$\cpm{0.00} & 1.00$\times$\cpm{0.00} & 1.00$\times$\cpm{0.00} & 1.00$\times$\cpm{0.00} & 28.95    \\
&$\text{Medusa}^{\blacklozenge}$                & 1.46$\times$\cpm{0.27} & 2.45$\times$\cpm{0.12} & 1.87$\times$\cpm{0.03} & 2.15$\times$\cpm{0.04} & 2.21$\times$\cpm{0.09} & 58.77  \\
&$\text{EAGLE}^{\blacklozenge}$ \cite{li2024eagle}                   & 2.43$\times$\cpm{0.05} & 3.16$\times$\cpm{0.07} & 2.58$\times$\cpm{0.02} & 2.85$\times$\cpm{0.06} & 2.77$\times$\cpm{0.1}  & 79.84   \\
&$\text{Hydra}^{\blacklozenge}$ \cite{ankner2024hydra}               & 1.79$\times$\cpm{0.26} & 3.11$\times$\cpm{0.12} & 2.2$\times$\cpm{0.03}  & 2.55$\times$\cpm{0.04} & \underline{2.81}$\times$\cpm{0.04} & 72.24  \\
\cmidrule[0.25pt](l){2-8}
&SpS \cite{chen2023accelerating}            & 1.75$\times$\cpm{0.06} & 1.97$\times$\cpm{0.08} & 2.04$\times$\cpm{0.06} & 1.78$\times$\cpm{0.05} & 1.73$\times$\cpm{0.06} & 53.58  \\
&Lookahead \cite{fu2024break}               & 1.16$\times$\cpm{0.24} & 1.7$\times$\cpm{0.07}  & 1.39$\times$\cpm{0.03} & 1.47$\times$\cpm{0.04} & 1.55$\times$\cpm{0.04} & 42.05  \\
&REST \cite{he2023rest}                     & 1.41$\times$\cpm{0.02} & 1.84$\times$\cpm{0.03} & 1.43$\times$\cpm{0.04} & 1.6$\times$\cpm{0.03}  & 1.69$\times$\cpm{0.02} & 46.25   \\
&PLD \cite{yang2023inference,saxena2023pld} & 2.62$\times$\cpm{0.03} & 2.43$\times$\cpm{0.04} & 2.73$\times$\cpm{0.02} & 3.11$\times$\cpm{0.06} & 1.63$\times$\cpm{0.02} & 72.48    \\
&PLD+ (a)                                   & 3.32$\times$\cpm{0.07} & 3.69$\times$\cpm{0.17} & 3.88$\times$\cpm{0.16} & 5.09$\times$\cpm{0.17} & 1.85$\times$\cpm{0.04} & 103.23  \\
&PLD+ (h)                                   & \textbf{\underline{3.39}$\times$\cpm{0.07}} & \textbf{\underline{3.83}$\times$\cpm{0.1}}  & \textbf{\underline{4.01}$\times$\cpm{0.01}} & \textbf{\underline{5.16}$\times$\cpm{0.1}} & \textbf{ 1.92$\times$\cpm{0.04}} & \underline{\textbf{106.05}}\\
\midrule
\multirow{11}{*}{\rotatebox{90}{\texttt{Vicuna-13B}}} 
&Autoregressive Decoding                    & 1.00$\times$\cpm{0.00} & 1.00$\times$\cpm{0.00} & 1.00$\times$\cpm{0.00} & 1.00$\times$\cpm{0.00} & 1.00$\times$\cpm{0.00} & 22.7   \\
& $\text{Medusa}^{\blacklozenge}$ \cite{cai2024medusa}                & 1.82$\times$\cpm{0.04} & 2.49$\times$\cpm{0.02} & 1.92$\times$\cpm{0.06} & 2.25$\times$\cpm{0.06} & 2.23$\times$\cpm{0.13} & 48.69  \\
&$\text{EAGLE}^{\blacklozenge}$ \cite{li2024eagle}                   & 2.51$\times$\cpm{0.04} & 3.32$\times$\cpm{0.14} & 2.73$\times$\cpm{0.05} & 3.03$\times$\cpm{0.07} & \underline{2.95}$\times$\cpm{0.03} & 66.08  \\
&$\text{Hydra}^{\blacklozenge}$\cite{ankner2024hydra}               & 2.24$\times$\cpm{0.09} & 3.01$\times$\cpm{0.46} & 2.37$\times$\cpm{0.09} & 2.82$\times$\cpm{0.04} & 2.89$\times$\cpm{0.05} & 60.62  \\
\cmidrule[0.25pt](l){2-8}
&SpS \cite{chen2023accelerating}            & 1.84$\times$\cpm{0.02} & 2.19$\times$\cpm{0.06} & 2.1$\times$\cpm{0.06}  & 1.78$\times$\cpm{0.08} & 1.74$\times$\cpm{0.07} & 43.81 \\
&Lookahead \cite{fu2024break}               & 1.39$\times$\cpm{0.02} & 1.68$\times$\cpm{0.09} & 1.33$\times$\cpm{0.02} & 1.44$\times$\cpm{0.04} & 1.52$\times$\cpm{0.04} & 33.49  \\
&REST \cite{he2023rest}                     & 1.44$\times$\cpm{0.01} & 1.95$\times$\cpm{0.05} & 1.41$\times$\cpm{0.04} & 1.74$\times$\cpm{0.06} & 1.7$\times$\cpm{0.02}  & 37.46  \\
&PLD \cite{yang2023inference,saxena2023pld} & 2.47$\times$\cpm{0.03} & 3.04$\times$\cpm{0.04} & 2.68$\times$\cpm{0.08} & 2.76$\times$\cpm{0.04} & 1.61$\times$\cpm{0.02} & 57.06 \\
&PLD+ (a)                                   & \textbf{\underline{2.75}$\times$\cpm{0.05}} & 5.2$\times$\cpm{0.18}  & 3.82$\times$\cpm{0.11} & 3.77$\times$\cpm{0.12} & 1.72$\times$\cpm{0.03} & 78.47  \\
&PLD+ (h)                                   & 2.73$\times$\cpm{0.02} & \textbf{\underline{5.37}$\times$\cpm{0.15}} & \textbf{\underline{3.88}$\times$\cpm{0.04}} & \textbf{\underline{3.85}$\times$\cpm{0.11}} & \textbf{1.79}$\times$\cpm{0.01} & \underline{\textbf{80.1 }} \\
\midrule
\multirow{10}{*}{\rotatebox{90}{\texttt{Vicuna-33B}}} 
&Autoregressive Decoding                    & 1.00$\times$\cpm{0.00} & 1.00$\times$\cpm{0.00} & 1.00$\times$\cpm{0.00} & 1.00$\times$\cpm{0.00} & 1.00$\times$\cpm{0.00} & 14.76   \\
&$\text{Medusa}^{\blacklozenge}$ \cite{cai2024medusa}  & 1.87$\times$\cpm{0.05} & 2.69$\times$\cpm{0.02} & 1.92$\times$\cpm{0.04} & 2.14$\times$\cpm{0.02} & 2.26$\times$\cpm{0.02} & 32.13 \\ 
&$\text{EAGLE}^{\blacklozenge}$ \cite{li2024eagle} & \underline{2.65}$\times$\cpm{0.05} & 3.74$\times$\cpm{0.11} & 2.85$\times$\cpm{0.09} & 3.04$\times$\cpm{0.09} & \underline{2.94}$\times$\cpm{0.05} & \underline{44.96} \\ 
&$\text{Hydra}^{\blacklozenge}$ & 2.33$\times$\cpm{0.05} & 3.41$\times$\cpm{0.11} & 2.32$\times$\cpm{0.05} & 2.8$\times$\cpm{0.02} & 2.92$\times$\cpm{0.06} & 40.7 \\ 
\cmidrule[0.25pt](l){2-8}
&SpS \cite{chen2023accelerating} & 1.87$\times$\cpm{0.05} & 2.42$\times$\cpm{0.05} & 2.34$\times$\cpm{0.12} & 1.87$\times$\cpm{0.01} & 1.8$\times$\cpm{0.03} & 30.38 \\ 
&Lookahead \cite{fu2024break} & 1.34$\times$\cpm{0.03} & 1.62$\times$\cpm{0.03} & 1.36$\times$\cpm{0.02} & 1.36$\times$\cpm{0.02} & 1.51$\times$\cpm{0.05} & 21.24 \\ 
&REST \cite{he2023rest} & 1.51$\times$\cpm{0.04} & 2.0$\times$\cpm{0.06} & 1.42$\times$\cpm{0.05} & 1.8$\times$\cpm{0.01} & 1.74$\times$\cpm{0.04} & 25.04 \\ 
&PLD \cite{yang2023inference,saxena2023pld} & \textbf{2.13}$\times$\cpm{0.0} & 2.77$\times$\cpm{0.04} & 2.87$\times$\cpm{0.22} & 2.45$\times$\cpm{0.03} & 1.56$\times$\cpm{0.02} & 34.75 \\ 
&PLD+ (a) & 2.07$\times$\cpm{0.04} & 3.71$\times$\cpm{0.07} & 4.2$\times$\cpm{0.04} & \underline{\textbf{3.22}}$\times$\cpm{0.01} & 1.59$\times$\cpm{0.06} & 43.65 \\ 
&PLD+ (h) & 2.09$\times$\cpm{0.02} & \underline{\textbf{3.8}}$\times$\cpm{0.11} & \underline{\textbf{4.29}}$\times$\cpm{0.1} & 3.18$\times$\cpm{0.03} & \textbf{1.59}$\times$\cpm{0.01} & \textbf{44.12} \\ 
\bottomrule
\end{tabular}}

\caption{\textbf{Comparison of PLD+ against various speculative decoding baselines across 5 input guided tasks (T=0).}  $\blacklozenge$ indicates tuning-dependent baselines. Mean speedup across 3 runs is reported. \textbf{Bold} represents best tuning-free and \underline{Underline} represents best overall. \textbf{Note:} Hyperparameters were chosen for PLD+ using the summarization task. }
\label{tab:main_exp}
\end{table*}

\begin{table*}[t!]
\small 
\centering
\resizebox{\linewidth}{!}{
\begin{tabular}{@{}ll|ccccc|cc@{}}
\toprule
\multicolumn{2}{l|}{\bf Methods} & \bf Summarization & \bf Code Editing & \bf Text Editing (Short) & \bf Text Editing (Long) &\begin{tabular}[c]{@{}c@{}}\bf Multi-turn\\\bf Conversation\end{tabular} &\begin{tabular}[c]{@{}c@{}}\bf Avg. Throughput\\\bf (\#tokens/s)\end{tabular}  \\
\midrule
\multirow{6}{*}{\rotatebox{90}{\texttt{Vicuna-7B}}} 
&Autoregressive Decoding                          & 1.00$\times$\cpm{0.00} & 1.00$\times$\cpm{0.00} & 1.00$\times$\cpm{0.00} & 1.00$\times$\cpm{0.00} & 1.00$\times$\cpm{0.00} & 27.02  \\
&$\text{EAGLE}^{\blacklozenge}$ \cite{li2024eagle} & 2.02$\times$\cpm{0.05} & \underline{2.75}$\times$\cpm{0.06} & 2.08$\times$\cpm{0.0}  & \underline{2.37$\times$\cpm{0.01}} & \underline{2.37}$\times$\cpm{0.16} & \underline{62.61} \\
\cmidrule[0.25pt](l){2-8}
&REST \cite{he2023rest}                           & 1.39$\times$\cpm{0.08} & 1.67$\times$\cpm{0.06} & 1.31$\times$\cpm{0.02} & 2.02$\times$\cpm{0.03} & \textbf{1.61}$\times$\cpm{0.1}  & 42.78  \\
&PLD \cite{yang2023inference,saxena2023pld}       & 2.04$\times$\cpm{0.1}  & 1.78$\times$\cpm{0.04} & 2.26$\times$\cpm{0.05} & 1.57$\times$\cpm{0.03} & 1.45$\times$\cpm{0.07} & 49.47 \\
&PLD+ (a)                                         & 2.28$\times$\cpm{0.09} & \textbf{2.52}$\times$\cpm{0.04} & \textbf{\underline{2.75}$\times$\cpm{0.06}} & 2.06$\times$\cpm{0.03} & 1.56$\times$\cpm{0.09} & 60.7 \\
&PLD+ (h)                                         & \textbf{\underline{2.33}$\times$\cpm{0.06}} & 2.2$\times$\cpm{0.02}  & 2.7$\times$\cpm{0.04}  & \textbf{2.26}$\times$\cpm{0.04} & 1.54$\times$\cpm{0.05} & \textbf{59.66} \\
\midrule
\multirow{6}{*}{\rotatebox{90}{\texttt{Vicuna-13B}}} 
&Autoregressive Decoding                          & 1.00$\times$\cpm{0.00} & 1.00$\times$\cpm{0.00} & 1.00$\times$\cpm{0.00} & 1.00$\times$\cpm{0.00} & 1.00$\times$\cpm{0.00} & 22.08 \\
&$\text{EAGLE}^{\blacklozenge}$ \cite{li2024eagle} & \underline{2.25}$\times$\cpm{0.04} & \underline{3.03}$\times$\cpm{0.14} & 2.32$\times$\cpm{0.02} & \underline{2.53}$\times$\cpm{0.09} & \underline{2.53}$\times$\cpm{0.07} & \underline{55.83 }\\
\cmidrule[0.25pt](l){2-8}
&REST \cite{he2023rest}                           & 1.4$\times$\cpm{0.02}  & 1.98$\times$\cpm{0.08} & 1.38$\times$\cpm{0.02} & 1.76$\times$\cpm{0.05} & \textbf{1.72}$\times$\cpm{0.05} & 36.32 \\
&PLD \cite{yang2023inference,saxena2023pld}       & 2.0$\times$\cpm{0.02}  & 2.02$\times$\cpm{0.1}  & 2.32$\times$\cpm{0.06} & 1.82$\times$\cpm{0.01} & 1.41$\times$\cpm{0.05} & 42.27 \\
&PLD+ (a)                                         & \textbf{2.24}$\times$\cpm{0.02} & 2.4$\times$\cpm{0.06}  & 3.16$\times$\cpm{0.05} & \textbf{2.21}$\times$\cpm{0.03} & 1.64$\times$\cpm{0.04} & \textbf{51.37} \\
&PLD+ (h)                                         & 2.09$\times$\cpm{0.03} & \textbf{2.53}$\times$\cpm{0.16} & \textbf{3.27}\textbf{$\times$\cpm{0.06}} & 2.15$\times$\cpm{0.08} & 1.54$\times$\cpm{0.01} & 51.01 \\
\midrule
\multirow{6}{*}{\rotatebox{90}{\texttt{Vicuna-33B}}} 
&Autoregressive Decoding                          & 1.00$\times$\cpm{0.00} & 1.00$\times$\cpm{0.00} & 1.00$\times$\cpm{0.00} & 1.00$\times$\cpm{0.00} & 1.00$\times$\cpm{0.00} & 14.41 \\
&$\text{EAGLE}^{\blacklozenge}$ \cite{li2024eagle} & \underline{2.36}$\times$\cpm{0.04} & 3.23$\times$\cpm{0.11} & 2.46$\times$\cpm{0.06} & \underline{2.67}$\times$\cpm{0.14} & \underline{2.63}$\times$\cpm{0.01} & \underline{38.66} \\
\cmidrule[0.25pt](l){2-8}
&REST \cite{he2023rest}                           & 1.51$\times$\cpm{0.03} & 1.99$\times$\cpm{0.04} & 1.46$\times$\cpm{0.04} & 1.84$\times$\cpm{0.07} & \textbf{1.68}$\times$\cpm{0.03} & 24.43 \\
&PLD \cite{yang2023inference,saxena2023pld}       & 1.96$\times$\cpm{0.06}  & 2.47$\times$\cpm{0.01} & 2.68$\times$\cpm{0.05} & 2.21$\times$\cpm{0.05} & 1.43$\times$\cpm{0.02} & 31.01 \\
&PLD+ (a)                                         & \textbf{1.97}$\times$\cpm{0.03} & \underline{\textbf{3.28}}$\times$\cpm{0.08} & 3.85$\times$\cpm{0.03}  & \textbf{2.57}$\times$\cpm{0.06} & 1.47$\times$\cpm{0.01} & \textbf{38.11} \\
&PLD+ (h)                                         & 1.89$\times$\cpm{0.03} & 3.14$\times$\cpm{0.08} & \underline{\textbf{3.86}}$\times$\cpm{0.1} & 2.39$\times$\cpm{0.06} & 1.52$\times$\cpm{0.02} & 36.91 \\
\bottomrule
\end{tabular}}
\caption{\textbf{Comparison of PLD+ against various speculative decoding baselines across 5 input guided tasks (T=1)} $\blacklozenge$ indicates tuning-dependent baselines.Mean speedup across 3 runs is reported. \textbf{Bold} represents best tuning-free and \underline{Underline} represents best overall. \textbf{Note:} Hyperparameters were chosen for PLD+ using the summarization task. }
\label{tab:sampling}
\end{table*}

\section{Experimental Setup}
\subsection{Datasets}
\label{sec:datasets}
For the tasks mentioned in Section \ref{sec:applications}, we sample data from the following datasets.

\textbf{Code Editing}: We leverage CodeEditorBench\_Plus, \footnote{The dataset was downloaded from \url{https://github.com/CodeEditorBench/CodeEditorBench}. It is licensed under Apache License 2.0
} one of the two datasets introduced by \cite{guo2024codeeditorbench} to test the performance of LLMs in code editing tasks: debugging, translating, polishing and requirement switching. We randomly select 20 instances for each task, resulting in a total of 80 samples.

\textbf{Text Editing (short)}: We leverage the text editing benchmark XATU \footnote{The dataset was downloaded from \url{https://github.com/megagonlabs/xatu}. It is licensed under CC-BY-NC 4.0 license}, introduced by \cite{zhang2023xatu} to test the capabilities of LLMS  for fine-grained instruction-based text editing. The benchmark accounts for tasks of various difficulties such as text simplification, grammar error correction, style transfer and information update.We randomly select 30 samples for each of the nine datasets resulting in a total of 270 samples.

\textbf{Text Editing (long)}: We leverage the ArgRewrite V.2 corpus\footnote{The dataset was downloaded from \url{https://argrewrite.cs.pitt.edu/\#corpus}. It is licensed under GNU General Public License} \cite{kashefi2022argrewrite} which contains annotated argumentative revisions, collected over 2 revision cycles on essays about self-driving cars. In particular, we make use of the first draft of the essay and the expert feedback (human feedback) given to the first draft. We sample 80 instances from this dataset to conduct our experiments.

\textbf{Multi-turn Conversation}: We leverage the MT-Bench benchmark \footnote{The dataset was downloaded from \url{https://huggingface.co/spaces/lmsys/mt-bench}. It is licensed under Apache License 2.0 } \cite{zheng2024judging} which consists of 80 multi-turn questions across the following categories: writing, roleplay, extraction, reasoning, math, coding, knowledge I (STEM), and knowledge II (humanities/social science). 

\textbf{Summarization }: We leverage the summarization subset of the Spec-Bench benchmark \footnote{The dataset was downloaded from \url{https://github.com/hemingkx/Spec-Bench}. It is licensed under Apache License 2.0}\cite{xia2024unlocking} for our experiments, as it has a high overlap between the input prompts and the generation outputs. This subset has 80 samples that were randomly sampled from the CNN/Daily Mail corpus \cite{nallapati-etal-2016-abstractive}.

\begin{figure*}[ht!]
    \centering
\includegraphics[width=\textwidth]{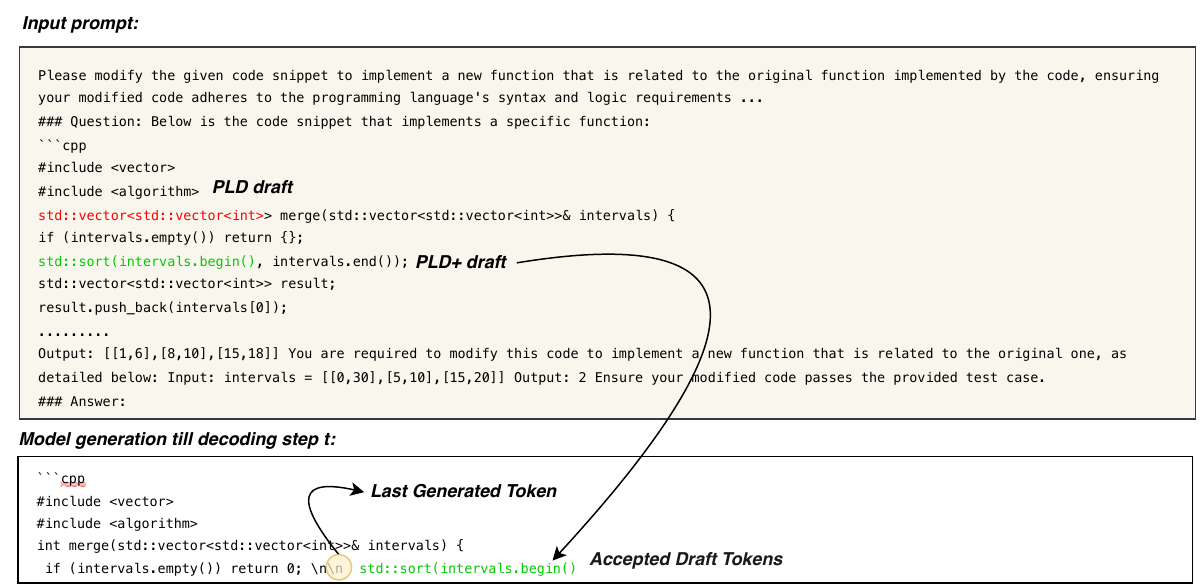}
    \caption{As shown in the figure, the last generated token is \string\n . PLD has proposed draft tokens by selecting the candidate span  with the longest matching prefix. However this is not always the optimum choice as shown in the figure.  PLD+ proposes the candidate span with the highest semantic relevance resulting draft token acceptance.}
    \label{fig:codepldvpldplus}
\end{figure*}
\subsection{Baselines}

We compare our method with seven speculative decoding approaches leveraging the Spec-Bench benchmark \cite{xia2024unlocking}. In particular we compare our approach with Medusa \cite{cai2024medusa}, EAGLE \cite{li2024eagle}, Hydra \cite{ankner2024hydra} (finetuning dependent approaches)
, and SpS \cite{chen2023accelerating}, Lookahead \cite{fu2024break}, LLM-A/PLD \cite{yang2023inference,saxena2023pld}, REST \cite{he2023rest} ( tuning-free approaches).



\subsection{Metrics}
Our evaluation centers on three key metrics: average throughput (tokens/sec), speedup (speculative vs. standard decoding), and average acceptance length (tokens accepted per generation).

\subsection{Implementation Details}
\label{sec:impl_det}
We conduct our main experiments using the \texttt{Vicuna-1.3} model series leveraging the code base of the Spec-Bench benchmark \cite{xia2024unlocking}. For SpS, we used \texttt{Vicuna-68m-v1.3} as the draft language model.  We follow the default parameters for all of our speculative decoding baselines. All our experiments have been conducted on an NVIDIA A100 GPU (with 80GB memory), with CUDA Version 12.2 and torch version 2.0.1.

\subsection{Hyperparameter Tuning}
\label{sec:hyperparam}

We used the summarization task as our validation set for hyperparameter tuning with \texttt{Vicuna-7b-1.3}. As explained in Section \ref{sec:pldplusdraft}, PLD+ has two key hyperparameters: the number of draft tokens (\textsc{K}) and the layer ($l$) if model hidden states are used to accelerate inference. We first determined the optimal value for $l$ by fixing \textsc{K}=10, then, for the selected layer, we tuned \textsc{K}. Using the average acceptance length metric, the best values were \textsc{K}=70 and $l$=9.

When using model attentions for accelerating inference, PLD+ has two hyperparameters: the number of draft tokens (\textsc{K}) and the attention heads ($g$). As described in Section \ref{sec:att_rank}, we identified 379 induction heads (37\% of the total heads in the model) and ranked them based on prefix matching and copying frequency. We found that using the top 50 heads gave the best average acceptance length. After fixing $g$ as the top-50 heads, we tuned \textsc{K} and found the optimal value to be \textsc{K}=70.

\section{Results and Analysis}
\label{sec:results}

\subsection{Main Experimental Results}

Tables \ref{tab:main_exp} (T=0, greedy decoding) and \ref{tab:sampling} (T=1, sampling) compare PLD+ with baselines across five input-guided tasks. We denote our approach as PLD+ (a) when using model attentions and PLD+ (h) when using hidden states.
In the greedy decoding scenario, PLD+ either exceeds or matches performance of tuning-free baselines in all five tasks. For code editing, short text editing, and long text editing tasks, PLD+ even surpasses the best-performing fine-tuned approach, EAGLE \cite{li2024eagle}, by margins ranging from $0.24\times$ to $2.85\times$ depending on model size. These results support our hypothesis that leveraging overlap between input and generation accelerates inference.

In the sampling scenario, speedups for all methods decrease compared to greedy decoding due to the randomness introduced by sampling. Despite this, PLD+ still outperforms all tuning-free baselines except on the Multi-turn Conversation task, where it ranks second, trailing REST \cite{he2023rest} by a maximum of 0.18. We manually reviewed the dataset and found that certain categories (roleplay, STEM, humanities) in the multi-turn conversation task may not qualify as input-guided tasks. The first-turn questions often lack sufficient information for generating second-turn answers, which impacts PLD+’s performance as it relies on input-output overlap.

Figure \ref{fig:codepldvpldplus} illustrates how PLD+ improves over PLD. In tasks like code editing, where repeated n-grams (e.g., indents) are common, correct ranking of drafts is crucial for performance. PLD+ surpasses PLD by ranking retrieved candidates based on semantic relevance using model artifacts rather than simple heuristics, yielding superior speedup.

While speculative decoding methods with powerful draft models or additional tuning could achieve even better alignment between the target LLM and the draft model, PLD+ offers an out-of-the-box, training-free solution for faster LLM inference, requiring no additional RAM for a draft model.

\begin{table}[t!]
\small 
\centering
\resizebox{\columnwidth}{!}{
\begin{tabular}{@{}ll|c|c@{}}
\toprule
\multicolumn{2}{l|}{\bf Methods} & \bf Speedup &  \bf Avg. Throughput  \\
\midrule
\multirow{5}{*}{\rotatebox{90}{\texttt{L2C-7B}}} 
 &Autoregressive Decoding & 1.00$\times$\cpm{0.00} & 28.73\cpm{0.26} \\
 &$\text{EAGLE}^{\blacklozenge}$ \cite{li2024eagle} & \underline{2.35}$\times$\cpm{0.04} & \underline{67.6}\cpm{0.88} \\ 
 \cmidrule[0.25pt](l){2-4}
 &REST \cite{he2023rest} & 1.48$\times$\cpm{0.01} & 42.45\cpm{0.38} \\
 &PLD \cite{yang2023inference,saxena2023pld} & 1.87$\times$\cpm{0.02} & 53.64\cpm{0.96} \\ 
&PLD+ (h) & \textbf{2.12}$\times$\cpm{0.04} & \textbf{61.01}\cpm{0.7} \\ 
\midrule
\multirow{5}{*}{\rotatebox{90}{\texttt{L2C-13B}}} 
&Autoregressive Decoding & 1.00$\times$\cpm{0.00} & 22.25\cpm{0.23} \\
&$\text{EAGLE}^{\blacklozenge}$ \cite{li2024eagle} & \underline{2.49}$\times$\cpm{0.03} & \underline{55.5}\cpm{0.3} \\
 \cmidrule[0.25pt](l){2-4}
&REST \cite{he2023rest} & 1.5$\times$\cpm{0.01} & 33.31\cpm{0.23} \\ 
&PLD \cite{yang2023inference,saxena2023pld} & 1.82$\times$\cpm{0.02} & 40.41\cpm{0.76} \\ 
&PLD+ (h) & \textbf{1.93}$\times$\cpm{0.05} & \textbf{42.89}\cpm{1.01} \\ 
\midrule
\multirow{3}{*}{\rotatebox{90}{\texttt{M-7B}}} 
&Autoregressive Decoding & 1.00$\times$\cpm{0.00} & 26.37\cpm{0.68} \\ 
&$\text{EAGLE}^{\blacklozenge}$ \cite{li2024eagle} & - & - \\ 
 \cmidrule[0.25pt](l){2-4}
&REST \cite{he2023rest} & - & - \\
&PLD \cite{yang2023inference,saxena2023pld} & 3.32$\times$\cpm{0.05} & 87.45\cpm{1.86} \\ 
&PLD+ (h) & \textbf{\underline{4.78}$\times$\cpm{0.02} }& \textbf{\underline{126.04}\cpm{3.59}} \\
\bottomrule
\end{tabular}}
\caption{Comparison of PLD+ with speculative decoding baselines on the summarization task using Mistral-7B-Instruct (M-7B), Llama-2-7B-Chat (L2C-7B)  and Llama-2-13B-Chat (L2C-7B) . Experiments were performed using with the greedy decoding strategy. Mean speedup across 3 runs is reported. \textbf{Bold} represents best tuning-free and \underline{Underline} represents best overall. }
\label{tab:new_models_main_table}
\end{table}

\subsection{Plug and play nature of PLD+}

To demonstrate the versatility of PLD+, we conducted experiments on the summarization task using newer models like Mistral-7B-Instruct, Llama-2-7B-Chat, and Llama-2-13B-Chat with the greedy decoding strategy. Unlike many speculative decoding methods that require additional compute/tuning and don’t support all LLMs out of the box, PLD+ works seamlessly across models. We compare against baselines that support these models, and as shown in Table \ref{tab:new_models_main_table}, PLD+ consistently outperforms all tuning-free baselines, reinforcing our claim that it can be used out of the box with any model, achieving notable speedups.

\begin{figure}
    \centering
    \includegraphics[width=\linewidth]{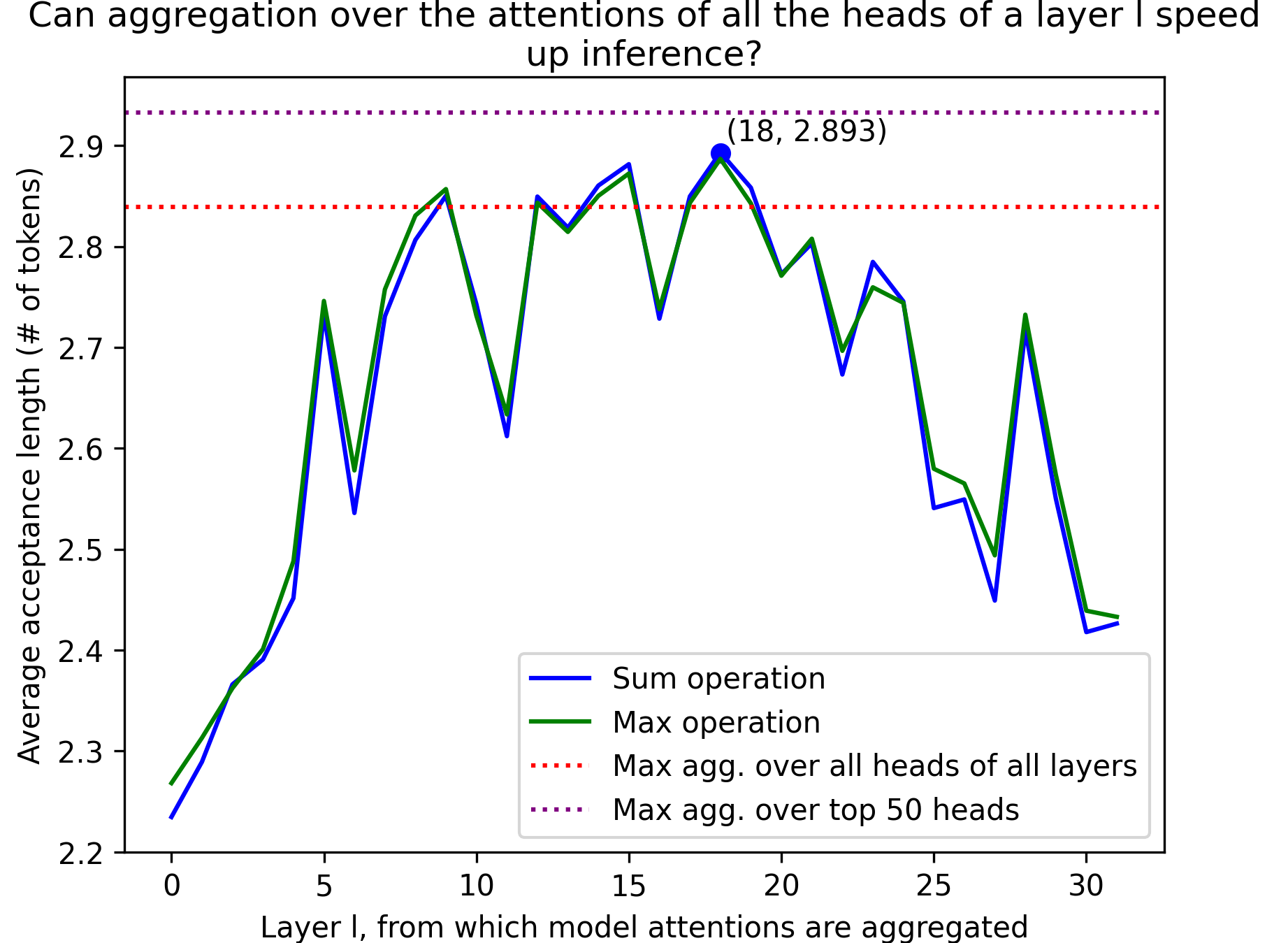}
    \caption{In this figure, we display the results of aggregating attentions across the heads of a specific layer. We experiment with two aggregation operations - summation and maximum aggregation.}
    \label{fig:att_heads}
\end{figure}

\section{Ablation study}

\subsection{How to choose attention heads for ranking occurrences?}
In Section \ref{sec:att_rank}, we detailed how we identify relevant attention heads. Using \texttt{Vicuna-7b-1.3} for the summarization task and average acceptance length as our metric, we tested other methods by aggregating attention from all heads in a given layer and across all layers, using both summation and maximum aggregation. Figure \ref{fig:att_heads} shows the results, with the best performance achieved by selecting specific attention heads (as described in Section \ref{sec:att_rank}) and applying maximum aggregation.

\subsection{Effect of number of draft tokens, \textsc{K} on performance of PLD+ and PLD}
\begin{figure}
    \centering
    \includegraphics[width=\columnwidth]{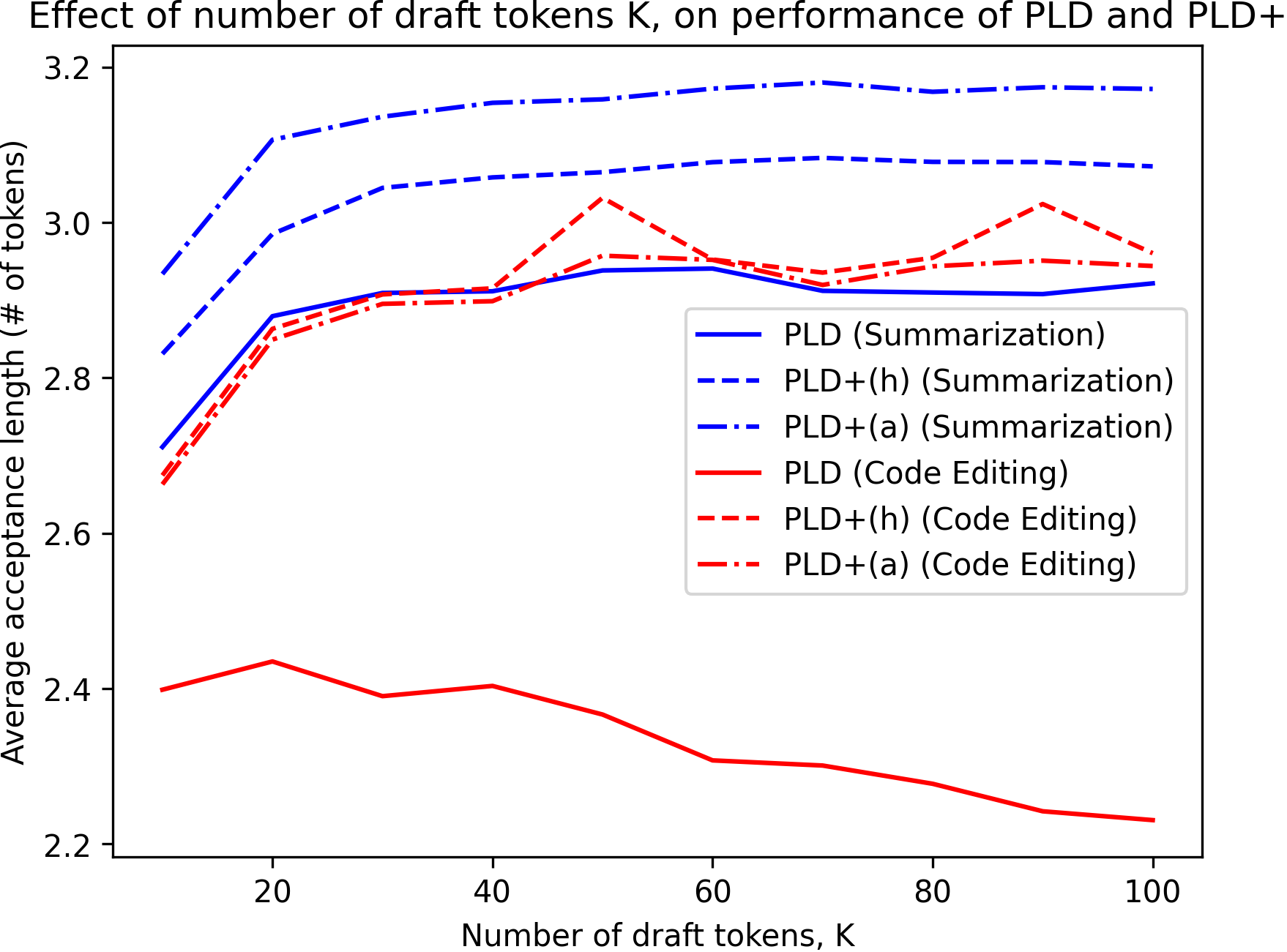}
    \caption{Effect of \textsc{K}, number of draft tokens on the performance of PLD, PLD+}
    \label{fig:Kvalue}
\end{figure}

In this section we analyse the impact of \textsc{K}, the number of draft tokens  on the performance of PLD and PLD+. We vary \textsc{K} from 10 to 100 and run experiments on the summarization task and use the average acceptance length metric to evaluate the performance. Figure \ref{fig:Kvalue} depicts the performance of PLD and PLD+ across different K values. From this figure, we can clearly see that PLD+ benefits from larger K values, and it would be more suitable for tasks where large spans of the output overlap with the input ( Code Editing, Text Editing). We hypothesize that using model artifacts (hidden states / attentions) to select the draft tokens helps us find longer matching spans compared to simple n-gram matching.

\subsection{Effect of layer l, on performance of PLD+(h)}
In this section we analyse the impact of the $l$ from which the hidden states are obtained for PLD+. For the \texttt{Vicuna-1.3} family of models, we fix \textsc{K}=10 and compute the performance for every $l$ on the summarization task and use the average acceptance length metric to evaluate the performance. Figure \ref{fig:Kvalue} depicts performance of PLD+(h) across different layers. From this figure, we can see that across models PLD+ benefits from using early layers (between 9 and 13).

\begin{figure}
    \centering
    \includegraphics[width=\columnwidth]{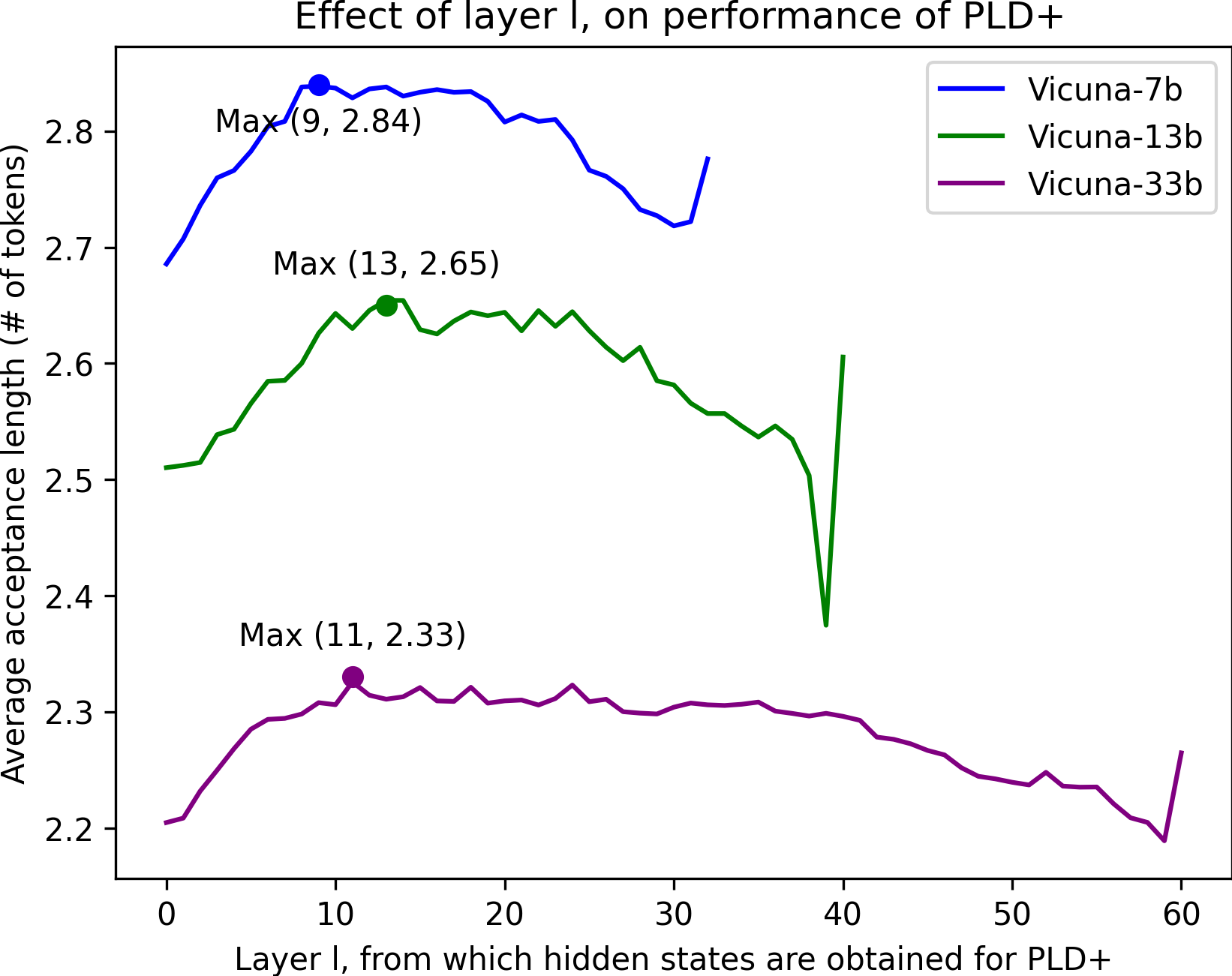}
    \caption{Effect of layer $l$ on performance of PLD+ (h) }
    \label{fig:layer}
\end{figure}

\section{Conclusion}
In this work, we propose Prompt Lookup Decoding+ (PLD+), a plug-and-play speculative decoding approach that can be easily integrated into the generation process of any language model without additional training. Instead of using a small language model as our drafter, we intelligently choose spans from the input as draft tokens. At each decoding step, we find candidate draft spans and choose the best span leveraging the model artifacts computed during generation. Through extensive experiments we found that PLD+ performs on par or even outperforms fine-tuning dependent speculative decoding approaches for the ubiquitous setting of input guided tasks.

\section{Limitations and Future Work}
Though PLD+ is a plug-and-play tuning-free decoding technique,  it is important to note that the speedup is directly influenced by the nature of the task. PLD+ is best suited for the tasks where there is a significant overlap between the input and the model generations. 

An important feature of PLD+ is the ability to point to the location in the context from which copying occurs. The task of identifying which parts of the context resulted in the generation of a token is known as input attribution and is of particular interest to the community \cite{li2023survey}.

In the Appendix \ref{sec:attribution}, we provide preliminary results on the ability of our method to perform attribution. Our approach to attribution has a similar flavor to the approach employed by \citet{phukan2024peering}, who successfully use LLM hidden states to perform attribution of verbatim copied spans. We intend to perform a detailed analysis of attribution quality in the future.

We want to highlight that our method focuses on speeding up inference and guarantees the same generation as standard autoregressive decoding.  Our method does not introduce any additional risks than those associated with LLMs such as bias, malicious use, etc.

\bibliography{custom}

\appendix
\begin{table*}[t!]
\small 
\centering
\resizebox{\linewidth}{!}{
\begin{tabular}{@{}ll|ccccc|c@{}}
\toprule
\multicolumn{2}{l|}{\bf Methods} & \bf Summarization & \bf Code Editing & \bf Text Editing (Short) & \bf Text Editing (Long) &\begin{tabular}[c]{@{}c@{}}\bf Multi-turn\\\bf Conversation\end{tabular} &\begin{tabular}[c]{@{}c@{}}\bf Avg. Throughput\\\bf (\#tokens/s)\end{tabular} \\
\midrule
\multirow{11}{*}{\rotatebox{90}{\texttt{Vicuna-7B}}} 
&Autoregressive Decoding & 1.00$\times$\cpm{0.00} & 1.00$\times$\cpm{0.00} & 1.00$\times$\cpm{0.00} & 1.00$\times$\cpm{0.00} & 1.00$\times$\cpm{0.00} & 28.53 \\ 
        &$\text{Medusa}^{\blacklozenge}$ \cite{cai2024medusa} & 1.8$\times$\cpm{0.06} & 2.51$\times$\cpm{0.06} & 1.78$\times$\cpm{0.06} & 2.2$\times$\cpm{0.04} & 2.22$\times$\cpm{0.01} & 59.97 \\ 
        &$\text{EAGLE}^{\blacklozenge}$ \cite{li2024eagle} & 2.48$\times$\cpm{0.08} & 3.19$\times$\cpm{0.05} & 2.51$\times$\cpm{0.05} & 2.84$\times$\cpm{0.05} & 2.74$\times$\cpm{0.03} & 78.59 \\ 
        &$\text{Hydra}^{\blacklozenge}$ \cite{ankner2024hydra} & 2.09$\times$\cpm{0.11} & 3.08$\times$\cpm{0.11} & 2.19$\times$\cpm{0.03} & 2.67$\times$\cpm{0.08} & 2.74$\times$\cpm{0.03} & 72.94 \\ 
        \cmidrule[0.25pt](l){2-8}
        &SpS \cite{chen2023accelerating} & 1.84$\times$\cpm{0.06} & 2.08$\times$\cpm{0.02} & 2.02$\times$\cpm{0.02} & 1.79$\times$\cpm{0.08} & 1.74$\times$\cpm{0.01} & 53.94 \\ 
        &Lookahead \cite{fu2024break} & 1.44$\times$\cpm{0.07} & 1.7$\times$\cpm{0.04} & 1.36$\times$\cpm{0.02} & 1.5$\times$\cpm{0.05} & 1.51$\times$\cpm{0.04} & 42.92 \\ 
        &REST \cite{he2023rest} & 1.41$\times$\cpm{0.02} & 1.83$\times$\cpm{0.03} & 1.38$\times$\cpm{0.01} & 1.65$\times$\cpm{0.03} & 1.61$\times$\cpm{0.04} & 45.09 \\ 
        &PLD \cite{yang2023inference,saxena2023pld} & 2.59$\times$\cpm{0.17} & 2.39$\times$\cpm{0.04} & 2.63$\times$\cpm{0.04} & 3.14$\times$\cpm{0.08} & 1.64$\times$\cpm{0.01} & 70.53 \\ 
        &PLD+ (a) & 3.43$\times$\cpm{0.09} & 3.7$\times$\cpm{0.19} & 3.84$\times$\cpm{0.02} & 5.37$\times$\cpm{0.12} & 1.89$\times$\cpm{0.04} & 103.9 \\ 
        &PLD+ (h) & 3.27$\times$\cpm{0.14} & 3.76$\times$\cpm{0.09} & 3.71$\times$\cpm{0.05} & 5.16$\times$\cpm{0.1} & 1.83$\times$\cpm{0.08} & \underline{\textbf{101.04}} \\
\bottomrule
\end{tabular}}

\caption{\textbf{Comparison of PLD+ against various speculative decoding baselines across 5 input guided tasks (T=0).} Experiments were performed using Vicuna-7b-v1.3 with the greedy decoding strategy. $\blacklozenge$ indicates tuning-dependent baselines. Mean speedup across 3 runs is reported. \textbf{Note:} Hyperparameters were chosen for PLD+ using the summarization task. }
\label{tab:40gbgreedy}
\end{table*}

\begin{table*}[t!]
\small 
\centering
\resizebox{\linewidth}{!}{
\begin{tabular}{@{}ll|ccccc|cc@{}}
\toprule
\multicolumn{2}{l|}{\bf Methods} & \bf Summarization & \bf Code Editing & \bf Text Editing (Short) & \bf Text Editing (Long) &\begin{tabular}[c]{@{}c@{}}\bf Multi-turn\\\bf Conversation\end{tabular} &\begin{tabular}[c]{@{}c@{}}\bf Avg. Throughput\\\bf (\#tokens/s)\end{tabular}  \\
\midrule
\multirow{6}{*}{\rotatebox{90}{\texttt{Vicuna-7B}}} 
 &Autoregressive Decoding & 1.00$\times$\cpm{0.00} & 1.00$\times$\cpm{0.00} & 1.00$\times$\cpm{0.00} & 1.00$\times$\cpm{0.00} & 1.00$\times$\cpm{0.00} & 28.45 \\ 
        &$\text{EAGLE}^{\blacklozenge}$ \cite{li2024eagle} & 2.07$\times$\cpm{0.04} & 2.63$\times$\cpm{0.02} & 2.05$\times$\cpm{0.04} & 2.34$\times$\cpm{0.08} & 2.29$\times$\cpm{0.05} & 64.77 \\ 
        \cmidrule[0.25pt](l){2-8}

        &REST \cite{he2023rest} & 1.43$\times$\cpm{0.04} & 1.78$\times$\cpm{0.05} & 1.34$\times$\cpm{0.02} & 1.69$\times$\cpm{0.04} & 1.63$\times$\cpm{0.04} & 44.81 \\ 
        &PLD \cite{yang2023inference,saxena2023pld} & 2.08$\times$\cpm{0.09} & 1.84$\times$\cpm{0.04} & 2.2$\times$\cpm{0.1} & 1.56$\times$\cpm{0.03} & 1.35$\times$\cpm{0.02} & 51.3 \\ 
        &PLD+ (a) & 2.32$\times$\cpm{0.11} & 2.57$\times$\cpm{0.04} & 2.67$\times$\cpm{0.05} & 1.68$\times$\cpm{0.03} & 1.57$\times$\cpm{0.07} & 61.44 \\ 
        &PLD+ (h) & 2.27$\times$\cpm{0.06} & 2.27$\times$\cpm{0.06} & 2.59$\times$\cpm{0.08} & 1.89$\times$\cpm{0.08} & 1.52$\times$\cpm{0.03} & 59.89 \\ 
\bottomrule
\end{tabular}}
\caption{\textbf{Comparison of PLD+ against various speculative decoding baselines across 5 input guided tasks (T=1)} $\blacklozenge$ indicates tuning-dependent baselines. Experiments were performed using Vicuna-7b-v1.3 with sampling and temperature=1. Mean speedup across 3 runs is reported. \textbf{Note:} Hyperparameters were chosen for PLD+ using the summarization task. }
\label{tab:40gbsampling}
\end{table*}

\section{Further analysis on A100 40GB GPU}
Using \texttt{Vicuna-7b-v1.3} as the LLM, we conduct all of our experiments on an NVIDIA A100 GPU (with 40GB memory), with CUDA Version 12.2 and torch version 2.0.1. Implementation details are mentioned in Section \ref{sec:impl_det}. Table \ref{tab:40gbgreedy} and Table \ref{tab:40gbsampling} indicate that PLD+ outperforms all of the tuning-free baselines on 4 input guided tasks, Code Editing, Summarization, Text editing (short), Text editing (long). In Table \ref{tab:40gbgreedy} PLD+ even outperforms all baselines in the greedy setting on the same four tasks.

\section{ Additional ablations}
\label{appendix:additional_ablations}
\subsection{How many top-k induction heads should we use for PLD+ (a)?}
Table \ref{tab:induction_head} summarizes the effect of the number of induction heads on the performance of PLD+(a).  We vary the number of top-k induction heads and perform experiments on the summarization dataset using \texttt{Vicuna-7b-v1.3} as the LLM. We can see that we get the best performance when the number of induction heads is 50.

\begin{table}
\centering
\resizebox{\columnwidth}{!}{%
\begin{tabular}{c|c}
 Number of top-k induction heads & Average Accept. Length \\
 \midrule
        10 & 3.17 \\ 
        20 & 3.18 \\ 
        30 & 3.2 \\ 
        40 & 3.21 \\ 
        50 & \textbf{3.21} \\ 
        100 & 2.93 \\ 
        200 & 2.9 \\ 
        300 & 2.9 \\ 
        All & 2.9 \\ 
\end{tabular}
}
\caption{Effect of the number of top-k induction heads on the performance of PLD+ (a) }
\label{tab:induction_head}
\end{table}

\subsection{Does using averaged hidden state representations improve performance of PLD+ (h)?}
In equation \ref{eq:hidden_state_computation}, averaged hidden state representations  $\overline{\mathbf{H}}^{(l)}$ can be used to find the best occurrence.  Let $m$ denote the prefix length used when computing the averaged representation, $\overline{\mathbf{H}}^{(l)}_{t} = \frac{1}{m} \sum_{i=1}^{m} \mathbf{{H}}^{(l)}_{i} $. We conduct experiments on the summarization task using \texttt{Vicuna-7B-v1.3} by varying the prefix length $m$. Table \ref{tab:avg} indicates that using averaged hidden state representations does not improve performance and that as $m$ increases performance decreases. In the generation setup, the hidden state of $x_{t+1}$ is contextualized by $x_t$ by design and $\mathbf{{H}}^{(l)}_{t+1}$ would already be similar to $\mathbf{{H}}^{(l)}_{t}$. We hypothesize that averaging the hidden states, $\mathbf{{H}}^{(l)}_{t}$ might lead to a less discriminative representation.

\subsection{Does thresholding improve performance of PLD+ (h)?}
In equation \ref{eq:hidden_state_computation}, we can additionally impose a threshold, where we consider an occurence $j$ only if the cosine similarity $\cos(\mathbf{H}^{(l)}_{j-1}, \mathbf{H}^{(l)}_{t-1})$ is greater than the threshold $\theta$. Setting the optimal $\theta$ value can help in reducing the number of occurences that need to be ranked in equation \ref{eq:hidden_state_computation}. We vary the value of $\theta$ from -1 to 1 and perform experiments on the summarization dataset using \texttt{Vicuna-7b-v1.3} as the LLM. Table \ref{tab:threshold} shows that $\theta=0.0$ gives the best performance. As $\theta>0.0$, the average throughput reduces indicating that the optimal draft span might have been discarded due to thresholding.

\begin{table}[]
\centering
\resizebox{\columnwidth}{!}{%
\begin{tabular}{cc|c}
&Experiment & Average Throughput (Tok/sec) \\
\toprule 
& PLD+, (No thresholding) & 92.26 \\
& PLD+, $\theta=-0.75$             & 94.44 \\
& PLD+, $\theta=-0.5$               & 93.79\\
& PLD+, $\theta=0$               & 95.05\\
& PLD+, $\theta=0.5$                 & 76.86 \\
& PLD+, $\theta=0.75$         & 57.73
\end{tabular}
}
\caption{Effect of threshold $\theta$ on performance of PLD+ (h)}
\label{tab:threshold}
\end{table}

\begin{table}[ht]
\centering
\resizebox{\columnwidth}{!}{%
\begin{tabular}{c|c}
Experiment  & Average Acceptance Length \\
\toprule
PLD+ (No averaging) & 3.14 \\
PLD+ ($m=2$)  & 3.05 \\
PLD+ ($m=3$)  & 3.04  \\
PLD+ ($m=4$)  & 3.03  \\
PLD+ ($m=5$) & 3.01\\
PLD+ ($m=6$) & 3.00 
\end{tabular}%
}
\caption{Effect of using averaged hidden state representations for PLD+ (h), where $m$ is the prefix length used for calculating the averaged representations. }  
\label{tab:avg} 
\end{table}

\begin{table}
    \centering
    \scriptsize
    \caption{Average number of words in the input for each input-guided task.}  
    \label{tab:input_word_count} 
    \resizebox{\columnwidth}{!}{\begin{tabular}{|l|l|}
        \hline
        Task (\# of samples) & Average \# of words in input \\ \hline
        Summarization (80) & 570.08 \\ \hline
        Code Editing  (80) & 289.19 \\ \hline
        \hspace{2em} debug  (20) & 317.55 \\ 
        \hspace{2em} polish (20) & 234.2 \\ 
        \hspace{2em} switch (20) & 289.9 \\ 
        \hspace{2em} translate (20) & 315.1 \\ \hline
        Text Editing (Short) (270) & 190.41 \\ \hline
        \hspace{2em} Information update (120) & 350.66 \\ 
        \hspace{2em} Style transfer (90) & 66.57 \\ 
        \hspace{2em} Simplification (30) & 52.87 \\ 
        \hspace{2em} Grammar correction (30) & 58.5 \\  \hline
        
        Text Editing (Long) (86) & 662.79 \\ \hline
        Multi-turn Conversation (80) & 34.15 \\ \hline
        \hspace{2em} writing (10) & 24.4 \\
        \hspace{2em} roleplay (10) & 30.3 \\ 
        \hspace{2em} reasoning  (10) & 34.85 \\ 
        \hspace{2em} math (10) & 23.05 \\ 
        \hspace{2em} coding (10) & 32.35 \\ 
        \hspace{2em} extraction (10) & 81.9 \\ 
        \hspace{2em} stem (10) & 22.85 \\ 
        \hspace{2em} humanities (10) & 23.5 \\ 
        \hline

    \end{tabular}}
    
\end{table}

\section{Dataset details}

In this section, we add additional details about the datasets used for each of the input-guided tasks. In Table \ref{tab:input_word_count}, we report the average number of words in the input to the LLM for each task. We offer a representative sample prompt for each category in the following subsections to illustrate the tasks.
\subsection{Summarization}
\begin{lstlisting}[basicstyle=\small\ttfamily, breaklines=true, breakatwhitespace=true]
Summarize: Hillary Clintons security detail arrived at a suburban Des Moines, Iowa fruit processing company on Tuesday with an added vehicle  a second Scooby. After her signature oversize black Chevy conversion van dropped her off at Capitol Fruit Company in Norwalk, Iowa, a visually identical GMC van drove up to the building with a nearly identical Secret Service escort vehicle. Both armored vehicles have raised roofs, deep-tinted windows and New York license plates...
\end{lstlisting}
\subsection{Code Editing}
\begin{lstlisting}[basicstyle=\small\ttfamily, breaklines=true, breakatwhitespace=true]
### Instruction:
Please correct the errors in the buggy code snippet below, ensuring that your corrected code adheres to the specified programming language syntax and logic requirements. Validate your solution against the provided test\ncases to ensure its accuracy. Note that your solution should strictly consist of the corrected code only. Generate only the required code enclosed in "```"
### Question:
Below is the java buggy code:
class Solution {\n    public int numberOfBeams(String[] bank) ...
Correct the code and ensure it passes the following test case:\nInput:  bank = [ "011001 ", "000000", "010100", "001000"] Output:  8
### Answer:
\end{lstlisting}

\subsection{Text Editing (Short)}
\begin{lstlisting}[basicstyle=\small\ttfamily, breaklines=true, breakatwhitespace=true]
Below is an instruction that describes a task, along with an input text. Please edit the input text based on the instruction. Your response should only include the edited output.
# Instruction:
Correct the grammar error in the sentence.
# Input:
When we talk about the so-called value of a product , we envision a scenario where dozens of products are available in supermarket shelves and when you switch on a television , there is an endless stream of commercials , each claiming exciting new features about the products advertised .
Response:
\end{lstlisting}
\subsection{Text Editing (Long)}
\begin{lstlisting}[basicstyle=\small\ttfamily, breaklines=true, breakatwhitespace=true]
    Essay: The article "Top 20 Pros and Cons Associated With Self-Driving Cars", provides an in depth look into the reasoning of whether or not we, as a society, should adopt self-driving vehicles.  Of the 20 "pros" listed, an increase in safety was likely the leading factor, but also included were the benefits in commute times and city travel, the possibility for the driver ...      
    Feedback:  Your draft has been read, and feedback from an expert writing instructor is written below. We advise that you use this feedback when you revise. 
    The strengths of your essay include:
    All of your sentences are clear because of word choice and sentence structure.
    You respond to one, but not all parts of the prompt. However, your entire essay is focused on the prompt.
    You provided specific and convincing evidence for each claim, and most evidence is given through relevant direct quotations or detailed examples from the provided reading.
    Areas to improve in your essay include: 
    You provided a statement that somewhat show your stance for or against self-driving cars, but it is unclear, or is just a restatement of the prompt.
    You made multiple, distinct, and clear claims that aligned with either your thesis, or the given reading, but not both. 
    Revise the essay using the expert feedback."
\end{lstlisting}
\subsection{Multi-turn Conversation}
\begin{lstlisting}[basicstyle=\small\ttfamily, breaklines=true, breakatwhitespace=true]
    TURN 1: Compose an engaging travel blog post about a recent trip to Hawaii, highlighting cultural experiences and must-see attractions.
    TURN 2: Rewrite your previous response. Start every sentence with the letter A.
\end{lstlisting}

\begin{table}[htbp]
\centering
\small
\begin{tabular}{|l|c|c|c|}
\hline
\textbf{Methods} & \textbf{Precision} & \textbf{Recall} & \textbf{F1} \\
\hline
GPT-4 & 0.96 & 0.87 & 0.90\\
Yi-6b \cite{phukan2024peering} & 0.94 & \textbf{0.99} & \textbf{0.96}\\
PLD+(h), threshold=2 & 0. 96 & 0.96 & \textbf{0.96}\\
PLD+(h), threshold=5 & 0.98 & 0.85 & 0.90\\
PLD+(h), threshold=8 & \textbf{0.99} & 0.7 & 0.79 \\
\hline
\end{tabular}
\caption{Token level P, R \& F1 scores for identifying output tokens copied from the context on QuoteSum test set}
\label{tab:attribution}
\end{table}

\begin{table*}[t!]
\small 
\centering
\resizebox{\linewidth}{!}{
\begin{tabular}{@{}ll|cccccc|c@{}}
\toprule
\multicolumn{2}{l|}{\bf Method} &\begin{tabular}[c]{@{}c@{}}\bf Multi-turn\\\bf Conversation\end{tabular} &\bf Translation & \bf Summarization & \begin{tabular}[c]{@{}c@{}}\bf Question\\\bf Answering\end{tabular} & \begin{tabular}[c]{@{}c@{}}\bf Mathematical\\\bf Reasoning\end{tabular} & \begin{tabular}[c]{@{}c@{}}\bf Retrieval-aug.\\\bf Generation\end{tabular} &  \bf Avg. \\
\midrule

\multirow{9}{*}{\rotatebox{90}{\texttt{Vicuna-7B}}} 
&Autoregressive Decoding &
  1.00$\times$\cpm{0.00} &
  1.00$\times$\cpm{0.00} &
  1.00$\times$\cpm{0.00} &
  1.00$\times$\cpm{0.00} &
  1.00$\times$\cpm{0.00} &
  1.00$\times$\cpm{0.00} &
  1.0 \\
&Medusa \cite{cai2024medusa} &
  2.21$\times$\cpm{0.08} &
  1.88$\times$\cpm{0.06} &
  1.79$\times$\cpm{0.07} &
  1.87$\times$\cpm{0.06} &
  2.19$\times$\cpm{0.07} &
  1.74$\times$\cpm{0.05} &
  1.95 \\
&EAGLE \cite{li2024eagle} &
  2.77$\times$\cpm{0.07} &
  2.02$\times$\cpm{0.05} &
  2.42$\times$\cpm{0.08} &
  2.2$\times$\cpm{0.06} &
  2.78$\times$\cpm{0.08} &
  2.23$\times$\cpm{0.07} &
  2.4 \\
&Hydra \cite{ankner2024hydra} &
  2.81$\times$\cpm{0.09} &
  2.23$\times$\cpm{0.05} &
  2.15$\times$\cpm{0.05} &
  2.27$\times$\cpm{0.06} &
  2.82$\times$\cpm{0.07} &
  2.14$\times$\cpm{0.06} &
  2.4 \\
&SpS \cite{chen2023accelerating} &
  1.73$\times$\cpm{0.06} &
  1.18$\times$\cpm{0.04} &
  1.77$\times$\cpm{0.06} &
  1.55$\times$\cpm{0.06} &
  1.52$\times$\cpm{0.05} &
  1.73$\times$\cpm{0.05} &
  1.58 \\
&Lookahead \cite{fu2024break} &
  1.55$\times$\cpm{0.04} &
  1.22$\times$\cpm{0.04} &
  1.45$\times$\cpm{0.05} &
  1.33$\times$\cpm{0.04} &
  1.67$\times$\cpm{0.05} &
  1.3$\times$\cpm{0.04} &
  1.42 \\
&REST \cite{he2023rest} &
  1.69$\times$\cpm{0.03} &
  1.34$\times$\cpm{0.04} &
  1.44$\times$\cpm{0.03} &
  1.75$\times$\cpm{0.03} &
  1.31$\times$\cpm{0.03} &
  1.62$\times$\cpm{0.04} &
  1.53 \\
&PLD \cite{yang2023inference,saxena2023pld} &
  1.63$\times$\cpm{0.06} &
  1.05$\times$\cpm{0.04} &
  2.62$\times$\cpm{0.09} &
  1.14$\times$\cpm{0.04} &
  1.61$\times$\cpm{0.06} &
  1.88$\times$\cpm{0.07} &
  1.66 \\
&PLD+ &
  1.83$\times$\cpm{0.05} &
  1.06$\times$\cpm{0.02} &
  3.23$\times$\cpm{0.08} &
  1.25$\times$\cpm{0.04} &
  1.77$\times$\cpm{0.04} &
  2.08$\times$\cpm{0.03} &
  1.87 \\

\midrule

\multirow{9}{*}{\rotatebox{90}{\texttt{Vicuna-13B}}} 
&Autoregressive Decoding &
  1.00$\times$\cpm{0.00} &
  1.00$\times$\cpm{0.00} &
  1.00$\times$\cpm{0.00} &
  1.00$\times$\cpm{0.00} &
  1.00$\times$\cpm{0.00} &
  1.00$\times$\cpm{0.00} &
  1.0 \\
&Medusa \cite{cai2024medusa} &
  2.23$\times$\cpm{0.12} &
  1.87$\times$\cpm{0.11} &
  1.83$\times$\cpm{0.05} &
  1.89$\times$\cpm{0.02} &
  2.31$\times$\cpm{0.03} &
  1.88$\times$\cpm{0.12} &
  2.0 \\
&EAGLE \cite{li2024eagle} &
  2.95$\times$\cpm{0.04} &
  2.16$\times$\cpm{0.05} &
  2.53$\times$\cpm{0.02} &
  2.25$\times$\cpm{0.02} &
  2.96$\times$\cpm{0.03} &
  2.48$\times$\cpm{0.15} &
  2.56 \\
&Hydra \cite{ankner2024hydra} &
  2.89$\times$\cpm{0.05} &
  2.23$\times$\cpm{0.03} &
  2.21$\times$\cpm{0.02} &
  2.32$\times$\cpm{0.03} &
  2.92$\times$\cpm{0.07} &
  2.31$\times$\cpm{0.16} &
  2.48 \\
&SpS \cite{chen2023accelerating} &
  1.75$\times$\cpm{0.07} &
  1.22$\times$\cpm{0.04} &
  1.82$\times$\cpm{0.06} &
  1.49$\times$\cpm{0.04} &
  1.64$\times$\cpm{0.06} &
  1.86$\times$\cpm{0.15} &
  1.63 \\
&Lookahead \cite{fu2024break} &
  1.52$\times$\cpm{0.03} &
  1.17$\times$\cpm{0.02} &
  1.38$\times$\cpm{0.01} &
  1.26$\times$\cpm{0.02} &
  1.7$\times$\cpm{0.04} &
  1.33$\times$\cpm{0.09} &
  1.39 \\
&REST \cite{he2023rest} &
  1.7$\times$\cpm{0.03} &
  1.31$\times$\cpm{0.0} &
  1.44$\times$\cpm{0.01} &
  1.7$\times$\cpm{0.04} &
  1.33$\times$\cpm{0.04} &
  1.69$\times$\cpm{0.12} &
  1.53 \\
&PLD \cite{yang2023inference,saxena2023pld} &
  1.61$\times$\cpm{0.04} &
  1.08$\times$\cpm{0.02} &
  2.46$\times$\cpm{0.06} &
  1.1$\times$\cpm{0.02} &
  1.7$\times$\cpm{0.05} &
  1.99$\times$\cpm{0.16} &
  1.66 \\
&PLD+ &
  1.74$\times$\cpm{0.03} &
  1.06$\times$\cpm{0.02} &
  2.66$\times$\cpm{0.04} &
  1.09$\times$\cpm{0.02} &
  1.79$\times$\cpm{0.04} &
  2.09$\times$\cpm{0.14} &
  1.74 \\
\midrule
\multirow{9}{*}{\rotatebox{90}{\texttt{Vicuna-33B}}}
&Autoregressive Decoding &
  1.00$\times$\cpm{0.00} &
  1.00$\times$\cpm{0.00} &
  1.00$\times$\cpm{0.00} &
  1.00$\times$\cpm{0.00} &
  1.00$\times$\cpm{0.00} &
  1.00$\times$\cpm{0.00} &
  1.0 \\
&Medusa \cite{cai2024medusa} &
  2.26$\times$\cpm{0.05} &
  1.97$\times$\cpm{0.03} &
  1.82$\times$\cpm{0.01} &
  1.88$\times$\cpm{0.01} &
  2.31$\times$\cpm{0.01} &
  1.8$\times$\cpm{0.01} &
  2.01 \\
&EAGLE \cite{li2024eagle} &
  2.94$\times$\cpm{0.05} &
  2.19$\times$\cpm{0.07} &
  2.56$\times$\cpm{0.05} &
  2.3$\times$\cpm{0.06} &
  3.15$\times$\cpm{0.08} &
  2.33$\times$\cpm{0.04} &
  2.58 \\
&Hydra \cite{ankner2024hydra} &
  2.92$\times$\cpm{0.06} &
  2.25$\times$\cpm{0.04} &
  2.23$\times$\cpm{0.03} &
  2.33$\times$\cpm{0.04} &
  2.94$\times$\cpm{0.07} &
  2.19$\times$\cpm{0.08} &
  2.48 \\
&SpS \cite{chen2023accelerating} &
  1.8$\times$\cpm{0.03} &
  1.29$\times$\cpm{0.01} &
  1.8$\times$\cpm{0.03} &
  1.54$\times$\cpm{0.02} &
  1.7$\times$\cpm{0.04} &
  1.68$\times$\cpm{0.03} &
  1.63 \\
&Lookahead \cite{fu2024break} &
  1.51$\times$\cpm{0.01} &
  1.21$\times$\cpm{0.03} &
  1.35$\times$\cpm{0.0} &
  1.31$\times$\cpm{0.01} &
  1.75$\times$\cpm{0.03} &
  1.29$\times$\cpm{0.02} &
  1.4 \\
&REST \cite{he2023rest} &
  1.74$\times$\cpm{0.02} &
  1.37$\times$\cpm{0.01} &
  1.47$\times$\cpm{0.02} &
  1.71$\times$\cpm{0.01} &
  1.35$\times$\cpm{0.03} &
  1.61$\times$\cpm{0.01} &
  1.54 \\
&PLD \cite{yang2023inference,saxena2023pld} &
  1.56$\times$\cpm{0.02} &
  1.1$\times$\cpm{0.01} &
  2.15$\times$\cpm{0.01} &
  1.11$\times$\cpm{0.0} &
  1.62$\times$\cpm{0.02} &
  1.54$\times$\cpm{0.0} &
  1.51 \\
&PLD+ &
  1.68$\times$\cpm{0.03} &
  1.09$\times$\cpm{0.03} &
  2.17$\times$\cpm{0.06} &
  1.1$\times$\cpm{0.06} &
  1.71$\times$\cpm{0.02} &
  1.55$\times$\cpm{0.01} &
  1.55 \\

\end{tabular}}
\caption{\textbf{Detailed comparison of PLD+ against various speculative decoding baselines on the Spec-Bench benchmark.} Experiments were performed using the \texttt{Vicuna-v1.3} series of models across different scales with the greedy decoding strategy. Mean speedup across 3 runs is reported.}
\label{tab:spec_benc_table}
\end{table*}

\section{Attribution}
\label{sec:attribution}
We utilise the ability of our method to point to the location in the context from where copying occurs to perform attribution. Following \citet{phukan2024peering}, who similar to us perform attribution of verbatim copied spans, we show results on the QuoteSum dataset \cite{schuster2023semqa} in Table \ref{tab:attribution}.

\label{sec:appendix}

\end{document}